\documentclass{article}

\PassOptionsToPackage{numbers,compress}{natbib}

\usepackage[preprint]{neurips_2026}

\usepackage[utf8]{inputenc}
\usepackage[T1]{fontenc}
\usepackage{microtype}

\usepackage{amsmath}
\usepackage{amssymb}
\usepackage{amsfonts}
\usepackage{nicefrac}
\usepackage{bm}

\usepackage{graphicx}
\usepackage{booktabs}
\usepackage{multirow}
\usepackage{makecell}
\usepackage{enumitem}
\usepackage{xcolor}

\usepackage{url}
\usepackage{hyperref}
\usepackage[capitalize]{cleveref}

\providecommand{\keywords}[1]{%
  \par\medskip\noindent
  {\renewcommand{\and}{\textperiodcentered\ }\textbf{Keywords:} #1}}

\title{Dynamic-in-Few-Step: Unifying Dynamic Computation and Few-Step Distillation for Efficient Video Generation}

\author{%
  Yu Cheng$^{1,2,3}$ \quad
  Siyue Yao$^{3,4}$ \quad
  Zhongang Qi$^{3,\dagger}$ \quad
  Shanyan Guan$^{3}$ \quad
  Wei Li$^{3}$ \quad
  Fajie Yuan$^{2,\dagger}$ \\[0.6em]
  {\small
   $^{1}$Zhejiang University \enspace
   $^{2}$Westlake University \enspace
   $^{3}$BlueImage, vivo \enspace
   $^{4}$Xian Jiaotong-Liverpool University} \\[0.4em]
  {
   \texttt{chengyu@westlake.edu.cn,\ qizhongang@vivo.com,\ yuanfajie@westlake.edu.cn}} \\[0.3em]
  {$^{\dagger}$Corresponding authors.}
}

\begin{document}

\maketitle

\begin{abstract}
Video Diffusion Models (VDMs) have demonstrated superior generation quality but suffer from prohibitive computational costs. While recent few-step distillation techniques significantly accelerate inference, they typically enforce a static model architecture across all denoising stages, ignoring the varying computational demands inherent to different noise levels. In this work, we propose a novel post-training acceleration framework that exploits this redundancy by integrating dynamic structural sparsification directly into the distillation process. Unlike conventional post-hoc compression applied to a fixed diffusion pipeline, our approach jointly optimizes the denoising steps and structured model sparsity, transforming a pre-trained VDM into a compact, step-specific Mixture-of-Models (MoM). To address the training instability arising from this joint optimization, we introduce a Progressive Training Strategy coupled with an Output Rollout Mechanism, which ensures the coherent learning of structural decisions across timesteps. Furthermore, we develop a specialized inference engine to deploy the resulting MoM efficiently. Our method is orthogonal to existing acceleration techniques and highly effective: On Wan-14B, it removes 24\% of the per-step FLOPs on top of 4-step distillation, adding a 1.2$\times$ wall-clock gain and reaching a 30$\times$ speedup over the 50-step teacher while preserving competitive generation quality.

\keywords{Video Generation Speedup \and Dynamic Network \and Efficient Diffusion}
\end{abstract}

\section{Introduction}
\label{sec:intro}

Video Diffusion Models (VDMs) \cite{wan2025wan, gao2025seedance, openai_sora, kong2024hunyuanvideo} have revolutionized video creation, enabling breakthroughs in simulation, advertising, and film production. However, their practical deployment is severely constrained by high inference latency and substantial computational resource requirements.

Recent advancements in post-training acceleration have been significant, focusing on step reduction \cite{zheng2025rcm, sun2025swiftvideo, lin2025diffusion}, attention optimization \cite{zhang2025fast, zhang2025sla}, quantization \cite{huang2025qvgen}, token compression \cite{chen2025dc}, and streamlined generation \cite{yin2025slow, huang2025self}. Despite notable progress, these methods share a common limitation: they treat the denoising network as a static architecture, applying the same heavy computation uniformly across all timesteps. However, the diffusion process is inherently heterogeneous. Existing works \cite{fan2025phased, zhao2024dynamic} observe that generation proceeds in a coarse-to-fine manner, where the model first constructs global structures and later refines high-frequency details. This suggests that distinct denoising stages require varying computational capacities, raising a pivotal question: \textit{Can we exploit this heterogeneity to dynamically allocate computation across timesteps, removing step-specific redundancy to accelerate VDMs?}

While prior works \cite{zhao2024dynamic, zhao2026dydit++} have explored dynamic computation, they are typically trained under standard denoising objectives and often become unstable or ineffective when adapted to few-step distillation \cite{zhao2024dynamic}. Given that few-step distillation has become the de facto paradigm for efficient VDMs, this incompatibility severely restricts their utility. Moreover, straightforward decoupled pipelines---either ``prune-then-distill'' or ``distill-then-prune''---are insufficient to resolve this mismatch. Specifically, pruning before distillation limits the model capacity required to learn the complex few-step mapping, while pruning after distillation disrupts the carefully aligned generation trajectory, leading to severe error accumulation, see \cref{sec:jointvssep}.

To translate dynamic computation into practical efficiency, we formulate dynamic pruning and few-step distillation as a unified optimization problem. Leveraging distribution matching---a robust objective validated in 3D generation \cite{wang2023prolificdreamer} and attention optimization \cite{huang2025self, zhang2025vsa}---we distill a dynamic structure specifically tailored to the compressed generation trajectory. This results in a step-aware Mixture-of-Models (MoM) framework capable of high-quality 4-step inference, where the network architecture is adaptively pruned at each step, inherently bypassing the incompatibility worries associated with post-hoc distillation.

Achieving this joint objective is non-trivial, as simultaneously optimizing for parameter sparsity and step reduction leads to gradient conflicts and training instability. To overcome this, we introduce a Progressive Training Strategy augmented with an Output Rollout mechanism. Our core insight is that diffusion is a sequential process where the ultimate objective is the perceptual quality of the final output. Guided by this, we adopt a reverse-order curriculum: this strategy first stabilizes the generation quality of the later denoising stages before progressively sparsifying the earlier, noisier stages. Furthermore, we leverage the output rollout strategy in Distribution Matching Distillation (DMD) \cite{yin2023onestep, yin2024improved}, enabling supervision based on the final output distribution. These designs ensure the stable training of highly efficient models. Finally, to translate theoretical sparsity into practical acceleration, we implement a specialized inference engine that effectively leverages this dynamic mixture of models.

Our contributions are summarized as follows:
\begin{enumerate}
\item We propose an end-to-end framework that jointly optimizes few-step distillation and dynamic structural sparsification, yielding a step-aware Mixture-of-Models that removes both temporal and parametric redundancy.
\item We introduce a progressive training plan combined with an output rollout strategy. This design effectively coordinates the co-optimization and ensures convergence to a high-quality solution.
\item Equipped with a specialized inference engine for the MoM architecture, our method removes 24\% of per-step FLOPs on Wan-14B and adds a 1.2$\times$ speedup over 4-step distillation, giving a 30$\times$ speedup over the 50-step teacher with competitive VBench scores.
\end{enumerate}

\section{Related Work}

\subsection{Efficient Video Diffusion Generation}
Acceleration techniques for pretrained Video Diffusion Models can be broadly categorized into training-free optimization \cite{liu2024timestep, lightx2v, dao2022flashattention, lu2022dpm} and post-training frameworks. We focus on the latter category in this work. Existing post-training approaches explore multiple dimensions of optimization. Step distillation methods \cite{zheng2025rcm, nie2026transition, fan2025phased, sun2025swiftvideo} compress the long sampling trajectory into a few-step regime (e.g., 4--8 steps), substantially reducing inference latency. Attention optimization techniques \cite{zhang2025vsa, zhang2025sla, huang2025linvideo} alleviate quadratic complexity through linearization or sparsification. Token reduction strategies \cite{chen2025dc} employ fine-tuning to adapt high-compression VAEs, effectively shortening spatial-temporal tokens. Causal distillation methods \cite{huang2025self, yin2025slow} convert slow bidirectional generation into efficient streaming generation. Quantization-aware training techniques \cite{huang2025qvgen} utilize low-bit parameterization to reduce inference overhead.

These approaches typically treat the denoising network as a static architecture, applying uniform computation across all timesteps. Our approach offers an orthogonal optimization perspective by targeting the structural redundancy inherent in the multi-step diffusion process, introducing step-dependent \textit{dynamic} optimization.

\subsection{Structural Pruning and Dynamic Networks}
Structural pruning removes redundant higher-level units (e.g., channels, filters, and blocks) to reduce parameters and FLOPs while preserving accuracy. Classical structural pruning identifies unimportant units based on magnitude~\cite{he2017channel,filters2016pruning}, importance scoring~\cite{molchanov2016pruning}, or gradient-sensitive criteria~\cite{theis2018faster}, and derives a compact static architecture for deployment. Dynamic networks extend this paradigm by introducing conditional computation, representative methods like \cite{lawson2025learning, han2024latency} dynamically adjust network depth or width based on input tokens. A prominent example is Mixture-of-Experts (MoE)~\cite{fedus2022switch, shazeer2017outrageously}, which activates subsets of parameters---typically in feed-forward layers---through learned routing mechanisms. While effective for scaling model capacity, these token-dependent dynamic structures impose strict requirements on routing strategies and system-level implementation to achieve practical wall-clock acceleration~\cite{fedus2022switch, dai2024deepseekmoe}.

In diffusion models, the varying noise levels across timesteps introduce a unique dimension of heterogeneity. DyDiT series \cite{zhao2024dynamic, zhao2026dydit++} explores this by proposing time-dependent pruning, learning a specific network structure for each timestep. As the routing depends only on the diffusion step, efficient batched inference is friendly supported without complex routing implementations.
However, DyDiT is trained under standard multi-step denoising objectives, lacking the mechanism to co-optimize with step reduction. As demonstrated in their original paper \cite{zhao2024dynamic}, combining it with step distillation leads to model collapse or ineffective training. This incompatibility with modern few-step distillation frameworks severely restricts its practical utility. We address this issue by designing a \textit{joint optimization} method.

\subsection{Step-Aware Mixture-of-Models}
Our method transforms a pretrained VDM into a step-aware Mixture-of-Models (MoM), where networks are pruned distinctly for different denoising stages. Several large-scale generative models, such as ERNIE-ViLG 2.0 \cite{feng2023ernie} and Wan2.2 \cite{wan2025}, also adopt MoM designs by training separate denoisers for different noise intervals. Their objective is to expand model capacity during pre-training to improve generation quality. Similarly, PhasedDMD \cite{fan2025phased} introduces a MoM structure within step distillation primarily for increased model expressiveness.

Conversely, we employ MoM as a post-training acceleration mechanism. We exploit the structural heterogeneity across noise levels to compress a pre-trained model, aiming for \textit{efficiency} rather than capacity expansion.

\section{Preliminaries}
\label{sec:preliminaries}

We perform step distillation based on the Distribution Matching Distillation method with specific modifications. Additionally, we apply the tri-level pruning scheme from \cite{wu2025taming} for structural sparsification. We briefly introduce these methods in this section.

\subsection{Modified Distribution Matching Distillation}
\label{sec:d_dmd}

We employ Decoupled Distribution Matching Distillation (D-DMD) \cite{liu2025decoupled} to distill a multi-step diffusion model into a $T$-step student generator $G_{\theta}$ (we set $T=4$). The student operates on a discrete set of timesteps $\mathcal{T} = \{t_1, t_2, \dots, t_T\}$, where $t_1$ denotes the initial highest noise level
and $t_T$ denotes the final lowest noise level prior to the clean data. The objective is to minimize the Kullback-Leibler (KL) divergence $\mathcal{D}_{\text{KL}}(p_{\theta} \| p_{\text{real}})$ between the student distribution $p_{\theta}$ and the real data distribution $p_{\text{real}}$. Since direct optimization of this divergence is intractable, DMD-based methods approximate the gradient $\nabla_{\theta} \mathcal{D}_{\text{KL}}$ with respect to the student parameters $\theta$ using score estimation (refer to \cite{yin2023onestep} for details). Specifically, D-DMD re-formulates the gradient by explicitly expanding the teacher's Classifier-Free Guidance (CFG) term and rearranging the components, resulting the following decoupled form:

\begin{equation}
\label{eq:dmd_gradient}
\resizebox{\linewidth}{!}{$\displaystyle
\nabla_{\theta} \mathcal{D}_{\scalebox{0.6}{\text{KL}}} \approx \mathbb{E}_{t \in \mathcal{T}, \tau_{\scalebox{0.6}{\text{DM}}}, \tau_{\scalebox{0.6}{\text{CA}}}} \left[ - \left( \underbrace{\left(\mathbf{s}_{\text{real}}(\mathbf{x}_{\tau_{\scalebox{0.6}{\text{DM}}}}, c) - \mathbf{s}_{\text{fake}}(\mathbf{x}_{\tau_{\scalebox{0.6}{\text{DM}}}}, c) \right)}_{\scalebox{0.9}{\text{Distribution Matching:}}\Delta_{\text{DM}}} + \underbrace{(\alpha-1) \left(\mathbf{s}_{\text{real}}(\mathbf{x}_{\tau_{\scalebox{0.6}{\text{CA}}}}, c) - \mathbf{s}_{\text{real}}(\mathbf{x}_{\tau_{\scalebox{0.6}{\text{CA}}}}, \emptyset)\right)}_{\scalebox{0.9}{\text{CFG Augmentation:}}\Delta_{\text{CA}}} \right) \frac{\partial G_{\theta}(\mathbf{z}_t)}{\partial \theta} \right],
$}
\end{equation}

where $\mathbf{z}_t$ denotes the latent input at few-step noise level $t \in \mathcal{T}$, and $\mathbf{x}_{\tau}$ represents the generator output re-noised to a continuous noise level $\tau$ ($\tau=0$ for pure noise, $\tau=1$ for clean data). The terms $c$ and $\emptyset$ denote the text condition and the null condition, respectively. $\mathbf{s}_{\text{real}}$ estimates the score of the real data distribution using the frozen pre-trained teacher network, and $\mathbf{s}_{\text{fake}}$ approximates the student's score via a fake network trained concurrently with a denoising objective on $G_{\theta}$'s generated samples.

The gradient comprises two components: $\Delta_{\text{CA}}$ encapsulates the CFG pattern ($\alpha$ is the CFG guidance scale), acting as the distillation engine that injects knowledge of teacher model to the student. $\Delta_{\text{DM}}$ serves as a regularizer, prevents the training process from diverging and ensures the quality of the final output. Following D-DMD, we apply decoupled re-noising schedules: $\tau_{\text{CA}} \sim \mathcal{U}[t, 1]$ focuses on refining fine-grained details in later denoising stages, whereas $\tau_{\text{DM}} \sim \mathcal{U}[0, 1]$ provides global regularization across all noise levels.

Moreover, we remove the ODE trajectory initialization in standard DMD as we found it causes generation instability. We incorporate the GAN loss suggested in DMD2 \cite{yin2024improved} to enhance video quality. Consequently, our modified distillation objective is formulated as $\mathcal{L}_{\text{DMD}} = \mathcal{D}_{\text{KL}} + \mathcal{L}_{\text{GAN}}$.

\subsection{Step-Aware Tri-level Structural Pruning}
\label{sec:prelim_pruning}

The Video Diffusion Model employed in our work is built upon the Diffusion Transformer (DiT) architecture.
Following previous practices~\cite{wu2025taming}, we adopt a tri-level pruning mechanism operating at the block, attention-head, and channel levels.
We formulate the pruning scheme using discrete binary gates $\mathbf{g} \in \{0, 1\}$, where $0$ indicates a pruned component and $1$ indicates a retained one.

To accommodate the sequential nature of few-step inference, we extend this static pruning into a step-aware dynamic structure.
Specifically, for each discrete inference step $t_i \in \mathcal{T}$,
we maintain a specific pruning scheme $\mathbf{g}_{t_i}$.
Consequently, the gate tensors for the block, head, and feed-forward network (FFN)
levels are defined with shapes $[T, N_{\text{block}}]$, $[T, N_{\text{head}}]$, and $[T, N_{\text{channel}}]$, where $N_{\text{block}}$, $N_{\text{head}}$, and $N_{\text{channel}}$ denote the total number of transformer blocks, multi-head attention heads, and FFN hidden channels, respectively.

\begin{figure}[t]
    \centering
    \includegraphics[width=\textwidth]{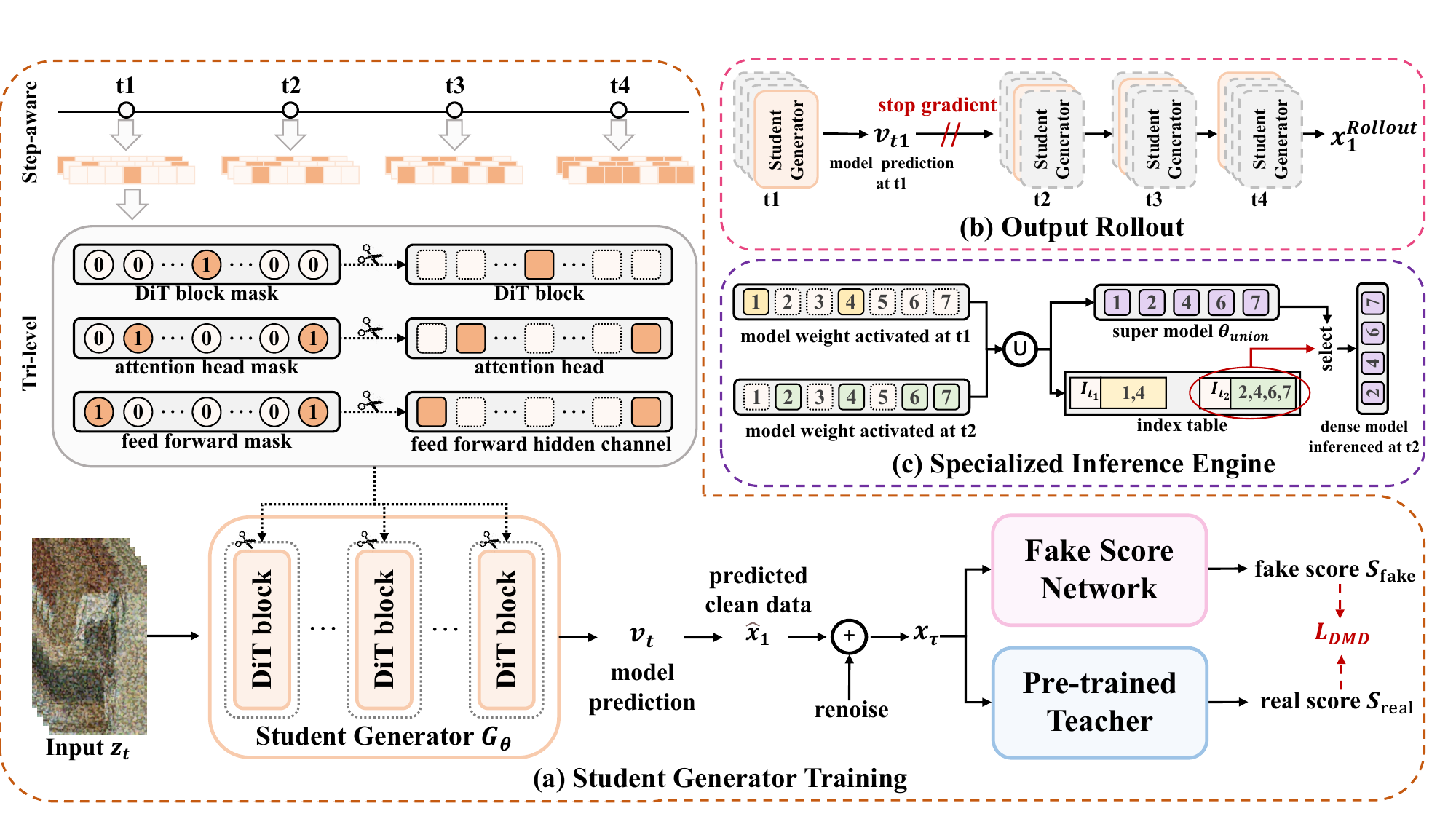}
    \caption{
        (a) \textbf{Unified training framework.} We jointly optimize few-step distribution matching and dynamic structural sparsification, learning step-adaptive architectures via mask-based gating. A single re-noising level $\tau$ is shown for simplicity.
       (b) \textbf{Output rollout mechanism.} Illustrated using $t_1$ as an example:
        starting from the intermediate noisy state at $t_1$, the student model is applied iteratively
        with detached gradients across all subsequent steps to produce the final rollout output
        $\mathbf{x}_1^{\text{rollout}}$, which is used to train the fake score network.
        (c) \textbf{Specialized inference engine.} Shown with $T{=}2$ steps and a model of 7 parameters for clarity.
        A unified super-model $\theta_{\text{union}}$ is constructed from the union of all active parameters,
        while step-specific index tables dynamically gather the required subsets during inference
        without redundant weight reloading.
    }
    \label{fig:framework}
\end{figure}

During training, these discrete gates are applied to the intermediate feature representations via element-wise multiplication to mask out redundant computations (see \cref{fig:framework}(a)).
Formally, for a specific architectural component $\mathcal{F}$ at inference step $t_i$,
the gated output is computed as:
\begin{equation}
\label{eq:gating}
    \mathbf{x}_{\text{out}} = \mathbf{g}_{t_i} \odot \mathcal{F}(\mathbf{x}_{\text{in}}),
\end{equation}
where $\odot$ denotes broadcasted element-wise multiplication. Once the gates are fully optimized, the resulting sparse architecture can be exported into a dense format.
As each inference step possesses a distinct structural configuration, the final exported model effectively constitutes a compact Mixture-of-Models (MoM), translating the theoretical sparsity of gates into actual computational speedup during inference.

\section{Method}
\label{sec:method}

In this section, we introduce our unified framework. Our goal is to \textit{learn a step-adaptive architecture that maximizes computational efficiency while preserving the high-fidelity generation capability of few-step distillation.} As illustrated in \cref{fig:framework}, our framework consists of three key components: (1) a joint optimization objective that unifies few-step distillation with dynamic structural pruning; (2) a progressive training strategy augmented with output rollout to stabilize the co-optimization process; and (3) a specialized inference engine designed to fully leverage the acceleration potential of the dynamic-in-few-step MoM architecture.

\subsection{Unifying Step Reduction and Dynamic Sparsification}
\label{sec:method_joint}

To exploit the temporal redundancy inherent in diffusion generation, we propose learning a dynamic pruning policy that adapts to the specific structural requirements of each denoising step.
Our strategy is to update the pruning gates directly via the few-step distillation objective. However, the discrete binary gates $\mathbf{g}_{t_i}$ defined in \cref{sec:prelim_pruning} are non-differentiable,
preventing standard gradient-based updates. To resolve this, we introduce continuous learnable structural masks $\mathbf{m}_{t_i}$ for each inference step $t_i \in \mathcal{T}$ to enable optimization within the distillation framework.

The mapping from the continuous soft masks $\mathbf{m}_{t_i}$ to the discrete hard gates $\mathbf{g}_{t_i}$ is tailored to each pruning granularity:
\begin{itemize}
 \item Block-level: the gate is directly determined by an indicator function $\mathbf{g}_{t_i} = \mathbb{I}(\mathbf{m}_{t_i} > 0)$.

 \item Attention-head level: for each head with a feature dimension of $D_{\text{head}}$, the entire head is retained ($\mathbf{g}_{t_i} = 1$) if the mean of its corresponding $D_{\text{head}}$ mask parameters is positive.

 \item FFN channel level: We group the hidden channels into sequential chunks of size $C$; a single scalar mask controls an entire chunk, yielding a block-wise gate vector $\mathbf{g}_{t_i} = \mathbb{I}(\mathbf{m}_{t_i} > 0) \cdot \mathbf{1}^C$.
\end{itemize}
We employ the Straight-Through Estimator (STE) to integrate the hard gates into the computational graph of the distillation task. Specifically, we construct a differentiable surrogate $\tilde{\mathbf{g}}_{t_i}$ for the forward pass:
\begin{equation}
\label{eq:ste}
    \tilde{\mathbf{g}}_{t_i} = \mathbf{g}_{t_i} + \mathbf{m}_{t_i} - \text{stop\_grad}(\mathbf{m}_{t_i}).
\end{equation}
During training, we substitute the discrete $\mathbf{g}_t$ with $\tilde{\mathbf{g}}_t$ in \cref{eq:gating}.
This formulation ensures that the forward pass utilizes the exact discrete gates to mask the corresponding components, while the backward pass routes the distillation task gradients directly to the continuous masks $\mathbf{m}_{t_i}$.

Furthermore, we introduce a sparsity penalty $\mathcal{L}_{\text{sparse}}$ to encourage structural sparsity and control the overall model capacity. When optimizing for step $t_i$, the global retention rate for component $k$ combines the differentiable surrogate $\tilde{\mathbf{g}}_{k, t_i}$ and the frozen hard gates $\mathbf{g}_{k, t_j}$ of other steps $t_j \neq t_i$:
\begin{equation}
\label{eq:sparse_loss}
    \mathcal{L}_{\text{sparse}}(\tilde{\mathbf{g}}_{t_i}) = \sum_{k \in \{b, h, f\}} \text{ReLU}\left( \frac{1}{T} \left( \mu(\tilde{\mathbf{g}}_{k, t_i}) + \sum_{t_j \neq t_i} \mu(\mathbf{g}_{k, t_j}) \right) - \eta_k \right),
\end{equation}
where $\mu(\cdot)$ calculates the spatial mean of the gate tensor for component $k$ (block, head, or FFN), and $\eta_{k}$ is the corresponding target sparsity ratio.

The overall training objective combines the distribution matching loss with the sparsity penalty:
\begin{equation}
    \mathcal{L}_{\text{total}} = \mathbb{E}_{t \sim \mathcal{T}} \left[ \mathcal{L}_{\text{DMD}}(\theta, \tilde{\mathbf{g}}_t)  + \lambda \mathcal{L}_{\text{sparse}}(\tilde{\mathbf{g}}_t) \right].
\end{equation}
By jointly optimizing the model parameters $\theta$ and the structural masks $\mathbf{m}$, our framework simultaneously uncovers both step and parameter redundancies. This co-optimization enables the model to learn specific structural demands across different timestep levels, leading to intriguing architectural findings discussed in \cref{sec:findings}.

\subsection{Progressive Training with Output Rollout}
\label{sec:method_progressive}

Directly optimizing the joint objective $\mathcal{L}_{\text{total}}$ results in unsatisfactory outcomes. We attribute this to the imbalanced distillation gradient norms across different noise levels. As the sparsity penalty is applied uniformly, stages with smaller distillation gradients are disproportionately dominated by the sparsity loss, leading to arbitrary over-pruning rather than learning genuinely step-adaptive structures.

To resolve this, we propose a progressive training strategy augmented with an output rollout mechanism. Our core insight is that diffusion is a sequential process where intermediate errors accumulate, yet the ultimate objective is solely the perceptual quality of the final output.
Therefore, we anchor our co-optimization to the final generation distribution.

First, we adopt a reverse-order curriculum to stabilize the structural search. We prioritize establishing a robust generation foundation at the later denoising stages (i.e., lower noise levels)
before progressively sparsifying the earlier, noisier stages. We initially optimize the distill loss and sparsity penalty exclusively for $t_4$ (the lowest noise level). Subsequently, we progressively incorporate earlier steps ($t_3 \rightarrow t_2 \rightarrow t_1$) into the training pipeline, and apply the sparsity penalty only to the newly introduced step. Formally, for the $T$-step model, the training proceeds backwards across $T$ progressive stages. In Stage $k$ ($k \in \{1, \dots, T\}$),
the active training subset is $\mathcal{T}_k = \{t_T, t_{T-1}, \dots, t_{T-k+1}\}$.
The newly introduced step is $t_{new} = t_{T-k+1}$. The objective for Stage $k$ is formulated as:
\begin{equation}
    \mathcal{L}_{\text{Stage-}k} = \mathbb{E}_{t \sim \mathcal{T}_k} \left[ \mathcal{L}_{\text{DMD}}(\theta, \tilde{\mathbf{g}}_t) \right] + \lambda \mathcal{L}_{\text{sparse}}(\tilde{\mathbf{g}}_{t_{new}}).
\end{equation}
After the $T$ progressive training stages, a final stage then jointly trains all $T$ steps using the full objective $\mathcal{L}_{\text{total}}$. This curriculum ensures the model secures its output quality before aggressively pruning the highly-noised input stages.

Furthermore, we align the training of the fake score network $\mathbf{s}_{\text{fake}}$ with this final-output-centric philosophy via an output rollout strategy.
In standard distillation practices, $\mathbf{s}_{\text{fake}}$ is typically trained on the single-step predicted clean data $\hat{\mathbf{x}}_1$ (computed from the model prediction ${\mathbf{v}}_t$) derived directly from an intermediate noisy state.
While effective for static architectures, this intermediate supervision exhibits reduced stability when dynamic structural pruning is introduced. Therefore, we choose to model the true multi-step generation outcome for a more reliable constraint.
Specifically, we employ an output rollout mechanism: when the student denoiser operates at an intermediate high-noise step (e.g., $t_1$),
we iteratively apply the student model---with gradients detached---to sequentially denoise the sample all the way to the final step $t_T$ (see \cref{fig:framework}(b)).
This final output ${\mathbf{x}}_1^{rollout}$ is then utilized as the training sample to update $\mathbf{s}_{\text{fake}}$.
By training the fake model to capture the final generation distribution, we inject a more robust regularization signal, ensuring the stable co-optimization of our dynamic MoM architecture.

\subsection{Specialized Inference Engine for MoM}
\label{sec:method_inference}
To translate the learned theoretical sparsity into tangible wall-clock acceleration, we design a specialized inference engine. After training, we construct a single unified \textit{super-model} containing the union of all active parameters across all steps: $\theta_{\text{union}} = \bigcup_{t \in \mathcal{T}} \{ \theta_i \mid m_{t, i} > 0 \}$.

Alongside this super-model, we maintain a lightweight index table $\mathcal{I}_t$ for each step $t \in \mathcal{T}$, explicitly pointing to the specific subset of $\theta_{\text{union}}$ required for that stage. During inference, the engine efficiently gathers the active dense parameters via $\mathcal{I}_t$. This dynamic routing eliminates the latency overhead of reloading different weights and significantly reduces memory consumption compared to storing $T$ separate models (see \cref{fig:framework}(c)).

\section{Experiments}
\label{sec:experiments}

\subsection{Implementation Details}

\textbf{Model Architecture and Initialization.}
We evaluate our method on Wan2.1-14B as the main model. We also use the smaller Wan2.1-1.3B for baseline comparison, ablation, and architectural analysis, where extensive training sweeps are tractable. Both models are trained for text-to-video generation at $480 \times 832$, generating 5-second videos at 16 frames per second (fps). The teacher and fake-score network are initialized from the full pre-trained checkpoint. The student generator adopts the same initialization but is augmented with our learnable structural masks (initialized to $1.0$) to enable dynamic pruning.

\noindent \textbf{Training Configurations.}
Our training subset comprises 50,000 high-quality videos curated from the Koala-36M dataset~\cite{wang2025koala}. We implement the discriminator architecture and GAN loss formulation following~\cite{chadebec2025flash}. The Classifier-Free Guidance scale is randomly sampled from $[3, 5]$ during training. We optimize the models using AdamW across 16 Kunlun P800 XPUs with a total batch size of 16. The learning rates are set to $1 \times 10^{-6}$ for the student model and fake score network, and $5 \times 10^{-7}$ for the GAN discriminator. The learning rate for the structural masks warms up from $1 \times 10^{-3}$ to a peak of $1 \times 10^{-2}$. The total training process spans 2,000 steps. For sparsity loss, the coefficient $\lambda$ is adaptively determined at each stage by monotonically mapping the corresponding gradient norm to the interval $(0,1)$. Additional implementation details are provided in the Appendix \cref{sec:app_training}.

\noindent \textbf{Progressive Pruning Schedule.}
We implement the reverse-order curriculum described in \cref{sec:method_progressive} using a FLOPs-driven stage transition criterion.
Specifically, we advance the training to incorporate the next noisier step (e.g., transitioning from $t_4$ to $t_3$) whenever the newly introduced step's FLOPs are reduced by $5\%$. We stop training once the average FLOPs across all steps $t \in \mathcal{T}$ drop to $75\%$ of the original dense model.

\subsection{Evaluation Metrics}
We evaluate our method along two dimensions: generation quality and computational efficiency. For generation quality, we adopt VBench~\cite{huang2024vbench} and report its eight primary dimensions, covering frame-wise quality (Imaging Quality, Aesthetic Quality), temporal quality (Motion Smoothness, Dynamic Degree, Background Consistency, Subject Consistency, Scene Consistency), and text alignment (Overall Consistency), following~\cite{huang2025linvideo}. The complete 16 metrics and additional qualitative results are provided in the Appendix \cref{sec:app_metrics}. We employ augmented prompts provided in VBench and generate five videos per prompt.
For efficiency, we report inference latency (seconds per video) measured on a single Kunlun P800 XPU. The reported runtime corresponds exclusively to the DiT backbone inference, excluding VAE decoding.

\begin{figure}[t]
    \centering
    \includegraphics[width=\linewidth]{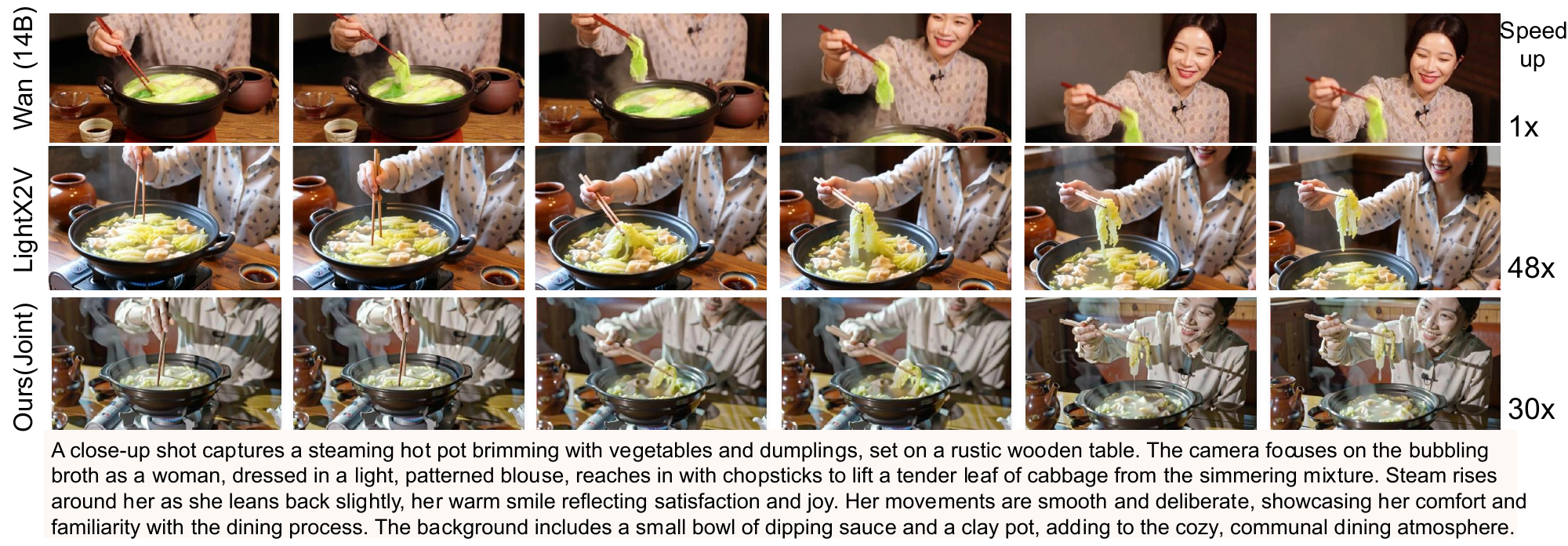}
    \caption{
        \textbf{Qualitative comparisons on Wan-14B.}
        We present our method with the 50-step Wan-14B
        teacher~\cite{wan2025wan} and the LightX2V acceleration
        framework~\cite{lightx2v}.
    }
    \label{fig:qualitative}
\end{figure}

\subsection{Baselines}
\label{sec:baselines}
Our method exposes a new acceleration axis---per-step computation redundancy.
Accordingly, our comparisons serve two purposes:
first, to test whether this step-wise pruning can be obtained by a two-stage pipeline;
second, to place our method in the context of existing acceleration frameworks.

\noindent \textbf{Two-Stage Optimization Baselines.} These baselines separate step distillation and structural pruning. They use the same 4-step distillation recipe as our method, but apply pruning either before or after distillation. For each setting, we sweep the pruning ratio and report the highest value that does not lead to visual collapse:

\noindent \textit{Category 1: Prune-then-Distill.} The model is first pruned and then distilled into a 4-step generator.

(1) Static $\rightarrow$ Distillation: We apply classical magnitude-based static pruning~\cite{he2017channel} to the FFN layers (attention head pruning is omitted as we observe it may trigger generation collapse). The sparsity ratio is capped at $10\%$.

(2) Dynamic $\rightarrow$ Distillation: Following DyDiT~\cite{zhao2024dynamic}, we train step-conditioned masks under the standard denoising objective with a sparsity loss. The noise schedule is divided into four intervals to match the later 4-step distillation. 

\noindent \textit{Category 2: Distill-then-Prune.} The model is first distilled into 4 steps and then pruned.

(3) Distillation $\rightarrow$ Static: The reverse pipeline of (1), where magnitude-based pruning is applied to a frozen distilled model.

(4) Distillation $\rightarrow$ Dynamic: We freeze the pre-distilled model weights and optimize only the dynamic masks using the same training pipeline described in \cref{sec:method}.

\noindent\textbf{Distillation Baseline and Teacher.}
\textit{Ours~(Distill)} is the 4-step DMD-distilled model without pruning, trained with the same data and recipe as \textit{Ours~(Joint)}. It is the direct baseline for measuring the extra benefit of our dynamic pruning. We also report the 50-step Wan model~\cite{wan2025wan} as the teacher and quality reference.

\noindent\textbf{SOTA Acceleration Frameworks.}
We report two representative open-source acceleration frameworks: TurboDiff~\cite{zhang2025turbodiffusion} (we evaluate the TurboWan2.1-T2V-1.3B checkpoint) and LightX2V~\cite{lightx2v} (the Wan2.1-T2V-14B-StepDistill-CfgDistill checkpoint). These methods combine step distillation with other optimizations, such as sparse attention, quantization, and inference-system acceleration. Their reported speedups (marked with $^{*}$ in \cref{tab:compare}) serve as a holistic reference for the overall acceleration landscape, but they are not direct comparison of our method: we study hidden-channel pruning, while these frameworks optimize other complementary dimensions. Because we cannot reproduce their full hardware and deployment environment, we cite their official latency gains; all VBench scores are measured under our own unified evaluation protocol.

\subsection{Main Results}
\label{sec:main_results}

\begin{table}[t]
    \centering
    \setlength{\tabcolsep}{2.6pt}
    \renewcommand{\arraystretch}{1.15}
    \caption{\textbf{Quantitative comparison on VBench.}
    Latency is measured on a single Kunlun P800 XPU; the speedup ratio is computed against the 50-step teacher \emph{within the same model scale}. ``$\Delta$ vs.\ Distill'' isolates the pruning contribution on top of 4-step distillation. Speedups marked with $^{*}$ are taken from the original publications as system-level references, rather than direct baselines.
    VBench scores for all methods are measured under our unified protocol. Best and second-best results within each scale are \textbf{bold}/\underline{underlined}.}
    \label{tab:compare}
    \resizebox{1.0\linewidth}{!}{
        \begin{tabular}{l | c c c | c c c c c c c c}
            \toprule
            Method &
            \makecell{Latency\\(s) $\downarrow$} &
            \makecell{Speedup\\vs.\ 50-step $\uparrow$} &
            \makecell{$\Delta$ vs.\\Distill $\uparrow$} &
            \makecell{Imaging\\Quality $\uparrow$} &
            \makecell{Aesthetic\\Quality $\uparrow$} &
            \makecell{Motion\\Smooth. $\uparrow$} &
            \makecell{Dynamic\\Degree $\uparrow$} &
            \makecell{Backgrnd.\\Consist. $\uparrow$} &
            \makecell{Subject\\Consist. $\uparrow$} &
            \makecell{Scene\\Consist. $\uparrow$} &
            \makecell{Overall\\Consist. $\uparrow$} \\
            \midrule
            \multicolumn{12}{c}{\textit{Wan-1.3B (5\,s, 16\,fps, $480\times832$)}} \\
            \midrule
            Wan (50 step)        & 325.84 & 1.00$\times$ & --     & 67.01 & 65.46 & 98.52 & 65.19 & \textbf{97.93} & \textbf{97.56} & \textbf{45.06} & 25.57 \\
            TurboDiffusion        & -- & 93.00$\times^{*}$ & 3.72$\times^{*}$     & 69.47 & 64.62 & 97.80 & \textbf{86.11} & 94.21 & 94.34 & 43.31 & 25.46  \\
            \midrule
            Static $\rightarrow$ Distill  & 12.92  & 25.22$\times$ & 1.01$\times$ & 69.05 & \underline{66.49} & \underline{98.72} & 56.94 & \underline{95.92} & \underline{97.19} & \textbf{45.06} & 25.57 \\
            Distill $\rightarrow$ Static  & 12.90  & 25.26$\times$ & 1.01$\times$ & 58.21 & 63.48 & 98.39 & 26.39 & 95.11 & 94.58 & 39.10 & 24.88 \\
            Dynamic $\rightarrow$ Distill & \underline{12.14}  & \underline{26.84}$\times$ & \underline{1.07}$\times$ & 67.57 & 64.37 & \textbf{98.74} & 65.28 & 95.68 & 95.31 & 40.84 & 25.24 \\
            Distill $\rightarrow$ Dynamic & 12.48  & 26.11$\times$ & 1.04$\times$ & 68.23 & 55.66 & 98.32 & 11.11 & 92.59 & 93.43 & 32.85 & 22.40 \\
            \midrule
            Ours (Distill)    & 12.93  & 25.20$\times$ & 1.00$\times$ & \underline{70.67} & \textbf{67.81} & 98.61 & 68.06 & 96.26 & 96.35 & \underline{44.11} & \textbf{25.88} \\
            \textbf{Ours (Joint)}         & \textbf{11.78} & \textbf{27.66}$\times$ & \textbf{1.09}$\times$ & \textbf{71.36} & 66.45 & 98.17 & \underline{80.56} & 95.63 & 95.80 & 43.68 & \underline{25.86} \\
            \midrule
            \multicolumn{12}{c}{\textit{Wan-14B (5\,s, 16\,fps, $480\times832$)}} \\
            \midrule
            Wan-14B (50 step)    & 1222.16 & 1.00$\times$  & --     & 69.43 & 66.07 & 98.30 & \textbf{65.46} & \textbf{98.09} & \textbf{97.52} & \underline{45.75} & \underline{25.91} \\
            LightX2V   & --     & 47.50$\times^{*}$         & 1.90 $\times^{*}$   & \textbf{71.15} & \underline{68.60} & \underline{98.45} & 58.80 & \underline{96.70} & 96.26 & \textbf{47.31} & \textbf{26.17} \\
            \textbf{Ours (Joint)} & \textbf{40.72} & \textbf{30.01}$\times$ & \textbf{1.20}$\times$ & \underline{69.31} & \textbf{69.36} & \textbf{98.90} & \underline{62.17} & 96.98 & \underline{97.15} & 45.57 & 26.12 \\
            \bottomrule
        \end{tabular}
    }
\end{table}

\cref{tab:compare} reports the results. We decompose each method's speedup into the step-reduction factor (relative to the 50-step teacher) and the additional pruning factor on top of 4-step distillation (``$\Delta$ vs.\ Distill''). Step distillation alone already gives a $\sim$25$\times$ saving; any further gain must come from making each remaining step cheaper, which the $\Delta$ column directly measures.

\noindent \textbf{Speedup on Wan-14B.}
On Wan-14B, Ours~(Joint) reduces per-step FLOPs by 24\% and cuts latency from 1222.16\,s (50-step teacher) to 40.72\,s. This represents a 30.01$\times$ end-to-end speedup over the teacher. Crucially, while step distillation contributes a 25.00$\times$ reduction, our dynamic pruning provides an additional 1.20$\times$ wall-clock acceleration. Despite this substantial capacity compression, our method maintains highly competitive VBench scores, confirming that the pruned capacity is genuinely redundant for the few-step trajectory.

\noindent \textbf{Decoupled pipelines fail to exploit per-step redundancy.}
\label{sec:jointvssep}
As shown in \cref{tab:compare}, decoupling pruning and distillation fails to deliver a favorable speed-quality trade-off.
First, \textit{Static$\rightarrow$Distill} and \textit{Distill$\rightarrow$Static} are limited by the static pruning pattern, yielding a negligible 1.01$\times$ marginal speedup.
Second, two-stage dynamic pipelines similarly falter due to sequential optimization conflicts. Applying pruning after distillation (\textit{Distill$\rightarrow$Dynamic}) severely disrupts the delicately aligned few-step generation trajectory, resulting in a precipitous degradation in temporal dynamics, most notably Dynamic Degree. Conversely, applying pruning prior to distillation (\textit{Dynamic$\rightarrow$Distill}) restricts the representational capacity essential for effectively learning the highly compressed 4-step mapping. Consequently, it suffers from noticeable performance drops in metrics like Imaging Quality and Dynamic Degree. Beyond quantitative metrics, our supplementary videos reveal that this variant generates visually inferior texture details and exhibits diminished motion amplitude.
In contrast, our unified framework resolves these capacity bottlenecks by jointly optimizing step reduction and dynamic pruning. This co-design achieves superior acceleration while preserving high-fidelity spatial details and rich, coherent temporal dynamics (further analysis in Appendix\cref{sec:app_dynamics} confirms the high dynamic degree is free from artifact-driven flickering).

\noindent \textbf{Comparison with SOTA frameworks.}
We position our method alongside integrated open-source acceleration frameworks, TurboDiffusion and LightX2V, to contextualize its performance. As demonstrated in \cref{tab:compare}, our method preserves competitive generation quality across VBench dimensions. It is important to emphasize that our approach is not designed to directly compete with these holistic, system-level pipelines, which bundle distillation with sparse attention, quantization, and infrastructure-level engineering. Instead, our contribution uncovers an orthogonal algorithmic acceleration axis: the per-step parametric redundancy along the hidden-channel dimension, driven by the coarse-to-fine dynamics of diffusion. Because our MoM exports standard dense architectures for each step, it is structurally compatible with these broader optimizations. We therefore position our method as complementary to, rather than competing with, such pipelines, with the potential to be combined for further gains.

\begin{figure}[t]
    \centering
    \includegraphics[width=\linewidth]{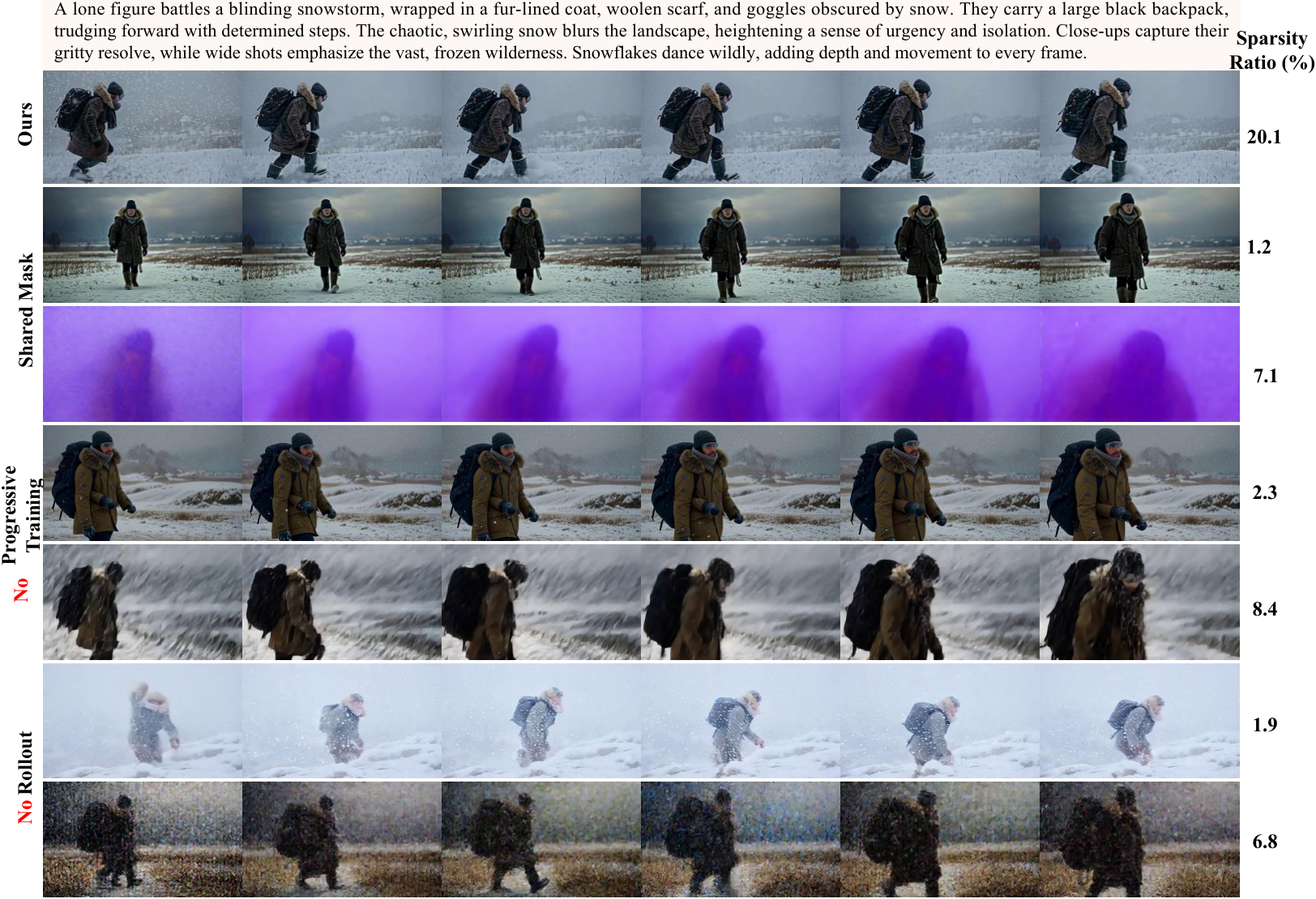}
    \caption{
        \textbf{Qualitative ablation of our training techniques.} The top row shows the stable generation of our full \textbf{Ours (Joint)} framework. The subsequent three groups represent the ablated settings. For each ablation, we display two states: the generation quality at its maximum viable sparsity (Upper), and the catastrophic collapse when pushed beyond this threshold (Lower). The sparsity ratio (calculated as pruned parameters over total parameters, averaged across 4 steps for step-aware models) is denoted on the right.
    }
    \label{fig:ablation}
\end{figure}

\subsection{Ablation Study}
\label{sec:ablation}

To validate the efficacy of our proposed training techniques, we conduct a comprehensive ablation study. Qualitative results are illustrated in \cref{fig:ablation}, with the corresponding quantitative bounds detailed in Appendix \cref{sec:app_quant_ablation}. Our experiments include:

\noindent \textbf{Step-Specific vs. Shared Mask.} We ablate the step-aware dynamic structure by enforcing a \textit{shared mask} across all inference steps.

\noindent \textbf{Effect of Progressive Training.} We remove the reverse-order curriculum, directly optimizing all steps jointly from the very beginning (i.e., exclusively using the Final Stage setup).

\noindent \textbf{Effect of Output Rollout.} We replace our output rollout mechanism and use the standard intermediate prediction $\hat{\mathbf{x}}_1$ for the fake score network training.

It can be observed that the step-specific configuration demonstrates superiority over the shared mask, as it generally achieves higher sparsity ratios and preserves basic structural fidelity. Furthermore, a consistent observation across all ablated settings is the existence of a strict sparsity wall. Without our tailored techniques, the models fail to reach high sparsity levels; forcing the optimization further leads to a cliff-like drop in structural integrity and a rapid collapse in generation quality. While the ablated models max out at a stable sparsity of only around 2.3\%, our full framework stably achieves a 20.1\% sparsity without visual degradation, highlighting the absolute indispensability of our training strategy for stabilizing the dynamic MoM co-optimization.

\section{Exploration and Findings}
\label{sec:findings}

\begin{figure}[t]
    \centering
    \includegraphics[width=\linewidth]{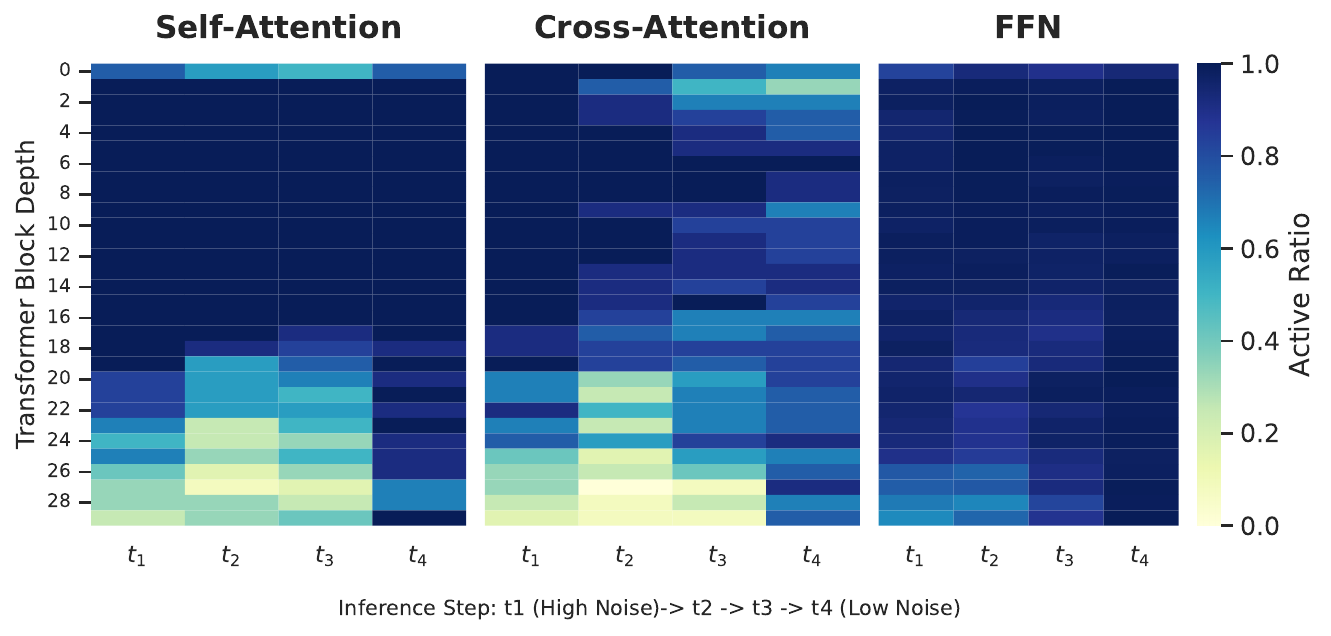}
    \caption{
        \textbf{Heatmap of the learned dynamic pruning policy.} We visualize the retention ratios of different architectural components (Self-Attention, Cross-Attention, and FFN) across different network depths and inference phases.
    }
    \label{fig:heatmap}
\end{figure}

By jointly optimizing the structural masks and the distillation objective, our framework automatically discovers how to allocate computation across denoising steps. Beyond the efficiency gains, the learned policy reveals interpretable structure in how redundancy is distributed both \emph{temporally} (across steps) and \emph{spatially} (across components and depth), as visualized in \cref{fig:heatmap} for the Wan-1.3B model. The Wan-14B model follows a similar pattern.

\textbf{U-Shaped Computational Allocation.}
Temporally, the learned architecture exhibits a distinct ``U-shaped'' capacity demand. The model retains the highest computational footprint at the two ends of the generation trajectory (88.4\% and 89.1\% FLOPs retention at $\mathbf{t}_{1}$ and $\mathbf{t}_{4}$, respectively) while aggressively pruning the intermediate steps (78.5\% and 77.1\% at $\mathbf{t}_{2}$ and $\mathbf{t}_{3}$). We further verified (\cref{fig:training_dynamics}(e)--(h)) that the U-shape emerges progressively during training and persists across a broad range of sparsity levels. The pattern mirrors the physics of diffusion generation: the initial phase needs ample capacity to construct the global layout from pure noise, the final phase to render high-frequency visual details, while intermediate phases mostly transition the established semantics and can be heavily compressed.

\textbf{Semantic Freezing in Deep Layers.}
Spatially, we observe that deeper layers are pruned much more aggressively than shallower layers, especially within the Self/Cross-Attention modules during $\mathbf{t}_{2}$ and $\mathbf{t}_{3}$. In Transformer-based diffusion models, deeper layers typically process high-level semantic representations, while shallower layers handle low-level textures. This pruning pattern suggests a semantic freezing phenomenon: high-level semantics are rapidly decided and frozen in the very early stages of inference. Consequently, the network automatically learns to shut down the deep-layer semantic updates in subsequent steps, allocating its limited computational budget to the shallow layers for fine-grained texture refinement.

\section{Conclusion}
\label{sec:conclusion}

We presented an end-to-end framework that unifies few-step distillation and dynamic structural pruning for accelerating Video Diffusion Models. By treating step reduction and parameter sparsification as a single joint optimization, we derive a step-aware Mixture-of-Models that allocates computation across denoising stages, and we keep this optimization stable with a reverse-order progressive curriculum and an output-rollout fake-score objective. On Wan-14B, our method achieves an 30$\times$ speedup while maintaining satisfied visual quality and temporal dynamics. We believe our unified framework and architectural findings will pave the way for designing fundamentally more efficient generative models in the future.

\bibliographystyle{plainnat}
\bibliography{main}

\newpage
\appendix
\renewcommand{\thesection}{\Alph{section}}
\begin{center}
\textbf{\Large Appendix}
\end{center}

This supplementary material provides further details and comprehensive results to support the main paper. It is organised as follows:
\begin{itemize}
    \item \textbf{Section~\ref{sec:app_training}} details the extended implementation hyperparameters and training configurations.
    \item \textbf{Section~\ref{sec:app_metrics}} presents the complete set of VBench evaluation metrics alongside additional qualitative visualisations.
    \item \textbf{Section~\ref{sec:app_quant_ablation}} provides the quantitative results for our ablation studies.
    \item \textbf{Section~\ref{sec:app_dynamics}} illustrates the training dynamics, the continuous quality-vs.-sparsity trade-off, and an in-depth analysis of our final model architecture.
\end{itemize}

\begin{figure}[htbp]
    \centering
    \includegraphics[width=\linewidth]{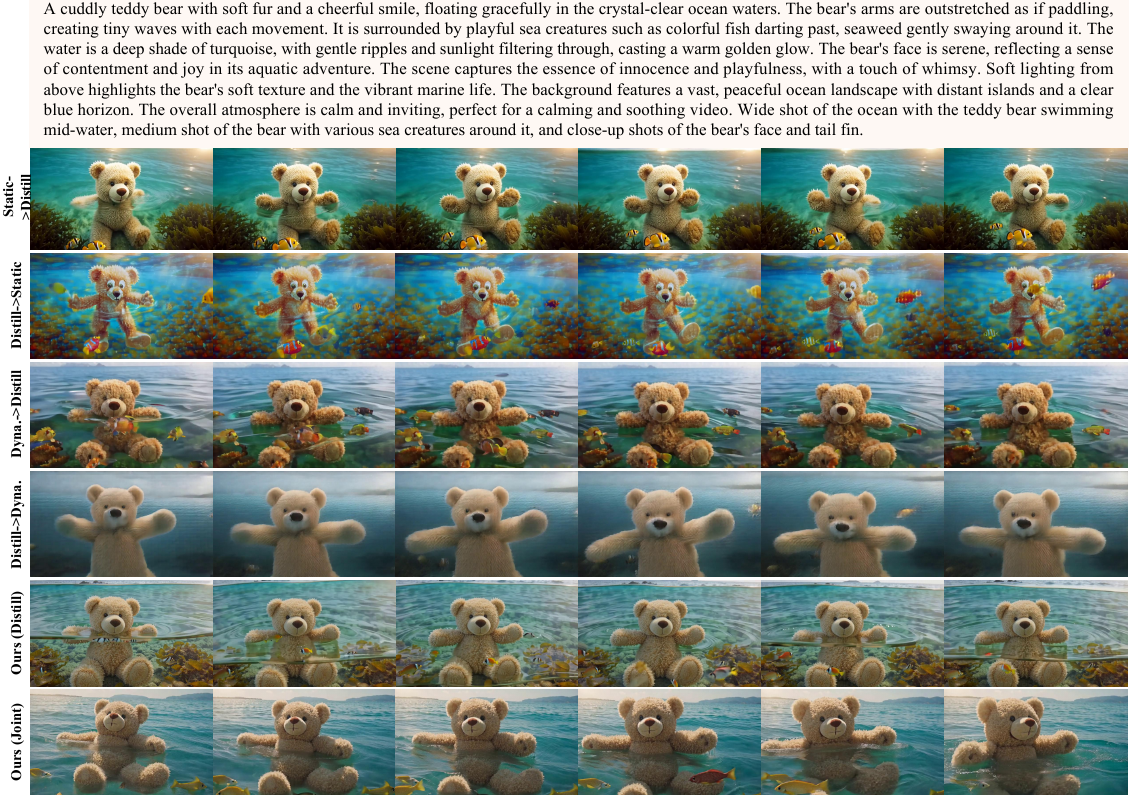}
    \caption{\textbf{Qualitative comparison with baselines.} Visual results demonstrating the generation quality of our proposed method compared against various baseline configurations. More example videos for these baselines are provided in the supplementary video material.}
    \label{fig:visual_baseline}
\end{figure}

\section{Extended Implementation Details}
\label{sec:app_training}

Due to space constraints in the main text, we provide the complete hyperparameter settings and training configurations here.

\textbf{Dynamic Sparsity Penalty Weighting.} As briefly discussed in the main text, we dynamically adjust the sparsity penalty weight $\lambda$ for each denoising step to balance the distillation and pruning objectives. Specifically, we maintain an Exponential Moving Average (EMA) of the task gradient norm, denoted as $n_t$, for each step $t$. To prevent the sparsity penalty from being overshadowed by large distillation gradients, we map this EMA value to a monotonically increasing scalar in the range $(0, 1)$ using the following smooth normalization function:
\begin{equation}
    \lambda_t = \frac{n_t + \epsilon}{n_t + \epsilon + 1},
\end{equation}
where $\epsilon = 10^{-6}$ is a small constant added for numerical stability.

\textbf{FFN Reordering and Chunking.} For the FFN channel-level pruning, we set the chunk size to $C=8$. We perform a pre-processing step before chunking: we reorder the FFN hidden channels of the pre-trained base model according to the $L_2$ norm of their corresponding weights. This norm-based sorting maximizes the intra-chunk homogeneity, allowing the single scalar mask to effectively control the entire chunk without sacrificing representational capacity.

\textbf{Memory Optimization and Hardware.} The model is trained using Fully Sharded Data Parallel (FSDP) combined with mixed-precision training (e.g., bfloat16). Additionally, we utilize gradient checkpointing to trade computation for memory, ensuring the joint optimization fits within the GPU memory limits. The training is conducted on 16 Kunlun P800 (96GB) XPUs.

\begin{table}[htbp]
\centering
\setlength{\tabcolsep}{2.6pt}
\renewcommand{\arraystretch}{1.15}
\caption{\textbf{Remaining VBench metrics on Wan 1.3 B.} The best and second-best results are highlighted in \textbf{bold} and \underline{underlined}, respectively. Visual comparisons are shown in Figure~\ref{fig:visual_baseline}.}
\label{tab:full_bench}
\resizebox{1.0\linewidth}{!}{
\begin{tabular}{l c c c c c c c c c c c}
\toprule
Baseline &
\makecell{Latency\\(s) $\downarrow$} &
\makecell{Speedup\\ratio $\uparrow$} &
\makecell{Object\\Class $\uparrow$} &
\makecell{Multiple\\Objects $\uparrow$} &
\makecell{Color $\uparrow$} &
\makecell{Spatial\\Relation. $\uparrow$} &
\makecell{Temporal\\Style $\uparrow$}  &
\makecell{Human\\Action $\uparrow$} &
\makecell{Temporal\\Flicker. $\uparrow$} &
\makecell{Appearance\\Style $\uparrow$} &
\makecell{\textbf{Total}\\\textbf{Score} $\uparrow$}\\
\midrule
Wan (50 step)  & 325.84 & 1.00$\times$ & \textbf{88.81} & 74.83 & \textbf{89.20} & 73.04 & 23.13 & 94.00 & \textbf{99.55} & \textbf{21.81} & 83.31 \\
\midrule
Static $\rightarrow$ Distill & 12.92 & 25.22$\times$ & 87.97 & \underline{80.26} & \underline{88.19} & \textbf{79.29} & 24.51 & \underline{95.00} & \underline{99.07} & 19.62 & 82.96 \\
Distill $\rightarrow$ Static & 12.90 & 25.26$\times$ & 80.70 & 69.89 & 76.44 & 75.67 & 23.58 & 93.00 & 98.52 & \underline{21.49} & 77.65 \\
Dynamic $\rightarrow$ Distill & \underline{12.14} & \underline{26.84}$\times$ & \underline{88.29} & 72.56 & 82.46 & 72.28 & 24.26 & 94.00 & 98.85 & 19.27 & 82.01 \\
Distill $\rightarrow$ Dynamic & 12.48 & 26.11$\times$ & 81.33 & 59.83 & 86.97 & 69.75 & 22.22 & 89.00 & 98.16 & 20.57 & 75.57 \\
\midrule
\textbf{Ours (Distill)} & 12.93 & 25.20$\times$ & 87.90 & \textbf{80.56} & 82.14 & 74.89 & \textbf{25.25} & \textbf{96.00} & 98.85 & 19.39 & \textbf{83.64} \\
\textbf{Ours (Joint)} & \textbf{11.78} & \textbf{27.66}$\times$ & 82.67 & 76.52 & 85.07 & \underline{75.87} & \underline{24.83} & \textbf{96.00} & 98.24 & 19.77 & \underline{83.62}\\
\bottomrule
\end{tabular}
}
\end{table}

\section{Full Evaluation Metrics and Additional Visualizations}
\label{sec:app_metrics}

The VBench suite comprehensively evaluates video generation models across multiple dimensions, including temporal quality (Subject Consistency, Background Consistency, Temporal Flickering, Motion Smoothness, Dynamic Degree), frame-wise quality (Imaging Quality, Aesthetic Quality), semantic consistency (Object Class, Multiple Objects, Human Action, Color, Spatial Relationship, Scene), style consistency (Appearance Style, Temporal Style), and text alignment (Overall Consistency).

In the main paper, we reported 8 core dimensions from the VBench suite. Here, we present the remaining 8 VBench metrics in Table~\ref{tab:full_bench}, along with the Total Score that reflects the comprehensive generation quality aggregated across all 16 dimensions.

\begin{figure}[htbp]
    \centering
    \includegraphics[width=\linewidth]{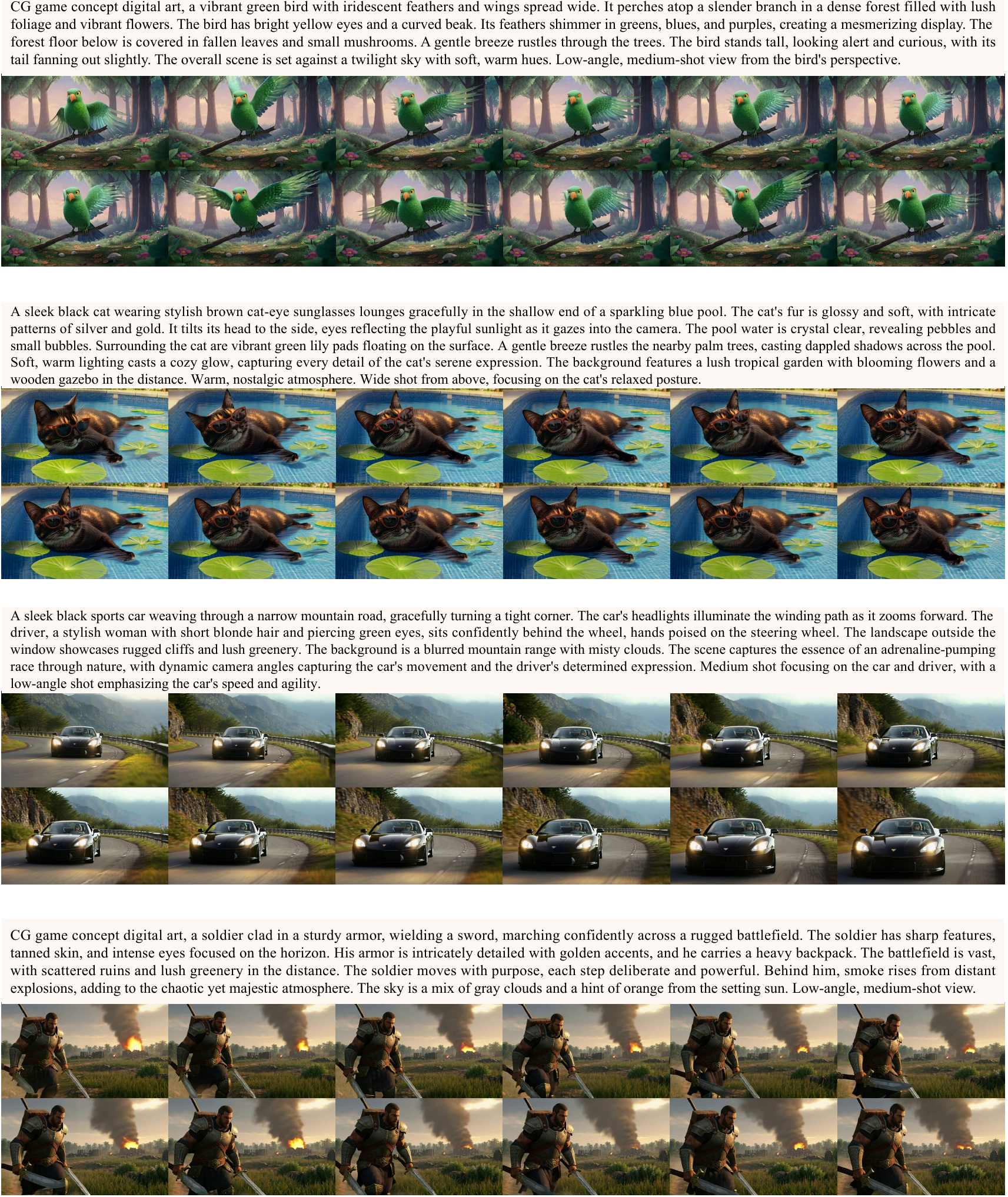}
    \caption{\textbf{Additional qualitative results of Ours (Joint).} The text prompts are randomly sampled from the VBench evaluation suite, showcasing the model's robustness across diverse scenarios without cherry-picking.}
    \label{fig:visual_mybaseline}
\end{figure}

\begin{figure}[htbp]
    \centering
    \includegraphics[width=\linewidth]{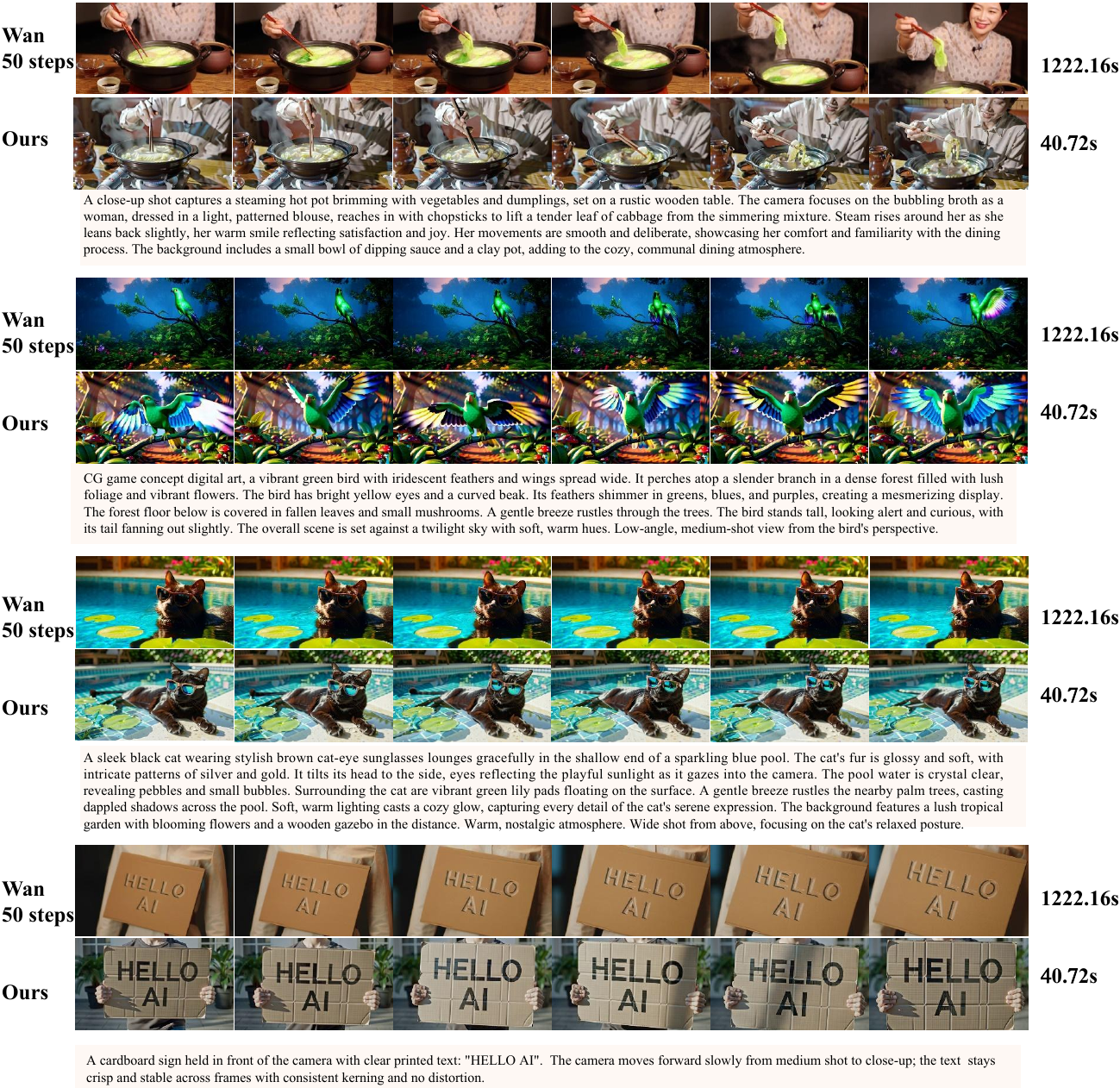}
    \caption{\textbf{Qualitative results on Wan 14B.} Visualisations demonstrate that our method maintains high generation quality with significant inference speedup on large-scale VDMs. Inference is conducted on a single P800 XPU using BF16 precision.}
    \label{fig:wan14b}
\end{figure}

As shown in Table~\ref{tab:full_bench}, consistent with the findings in the main text, our joint optimization method (\textbf{Ours (Joint)}) achieves a highly competitive Total Score of 83.62, closely matching our dense distillation baseline (\textbf{Ours (Distill)}, 83.64) while reaching a 27.66$\times$ end-to-end speedup over the 50-step teacher on Wan-1.3B. Additional qualitative results of \textbf{Ours (Joint)} are provided in Figure~\ref{fig:visual_mybaseline} and  Figure ~\ref{fig:wan14b}. 

\section{Quantitative Ablation}
\label{sec:app_quant_ablation}
Ablated variants of our dynamic MoM framework experience rapid generation collapse before reaching meaningful sparsity ratios. \cref{tab:ablation} provides the quantitative metrics of these failures on Wan-1.3B.

The results clearly demonstrate that without our proposed progressive training and output rollout mechanisms, the optimization hits a rigid ``sparsity wall'' at roughly 2.0\%--2.5\%. Pushing beyond this threshold results in collapsed models that suffer severe degradation across core metrics, particularly in Imaging Quality, Aesthetic Quality, and Dynamic Degree. While our main experiments demonstrate that joint training is the key to unlocking dynamic acceleration, the ablation study reveals that such co-optimization is highly non-trivial. It highlights the absolute indispensability of our progressive curriculum and rollout strategies for stabilizing the dynamic MoM co-optimization.

\begin{table}[htbp]
    \centering
    \setlength{\tabcolsep}{3pt}
    \renewcommand{\arraystretch}{1.10}
    \caption{\textbf{Quantitative ablation on Wan-1.3B.} These are the collapsed results beyond the $\sim$2.3\% stability wall.}
    \label{tab:ablation}
    \resizebox{\linewidth}{!}{
    \begin{tabular}{l c | c c c c c c c c}
        \toprule
        Variant & Sparsity &
        \makecell{Img.\\Q. $\uparrow$} & \makecell{Aes.\\Q. $\uparrow$} & \makecell{Mot.\\Sm. $\uparrow$} & \makecell{Dyn.\\Deg. $\uparrow$} & \makecell{Bg.\\Cons. $\uparrow$} & \makecell{Subj.\\Cons. $\uparrow$} & \makecell{Scene\\Cons. $\uparrow$} & \makecell{Over.\\Cons. $\uparrow$} \\
        \midrule
        \textbf{Ours (Joint)}      & 20.1\% & 71.36 & 66.45 & 98.17 & 80.56 & 95.63 & 95.80 & 43.68 & 25.86 \\
        \midrule
        (A) Shared Mask                  & 7.1\%  & 56.19 & 49.96 & 97.43 & 56.67 & 91.55 & 84.71 & 27.19 & 23.27 \\
        \midrule
        (B) w/o Progressive Training     & 8.4\%  & 52.41 & 47.83 & 97.18 & 36.94 & 90.27 & 80.49 & 24.66 & 22.75 \\
        \midrule
        (C) w/o Output Rollout           & 6.8\%  & 54.88 & 49.12 & 97.36 & 41.22 & 91.04 & 82.36 & 25.31 & 23.04 \\
        \bottomrule
    \end{tabular}}
\end{table}

\begin{figure}[t]
    \centering
    \includegraphics[width=\linewidth]{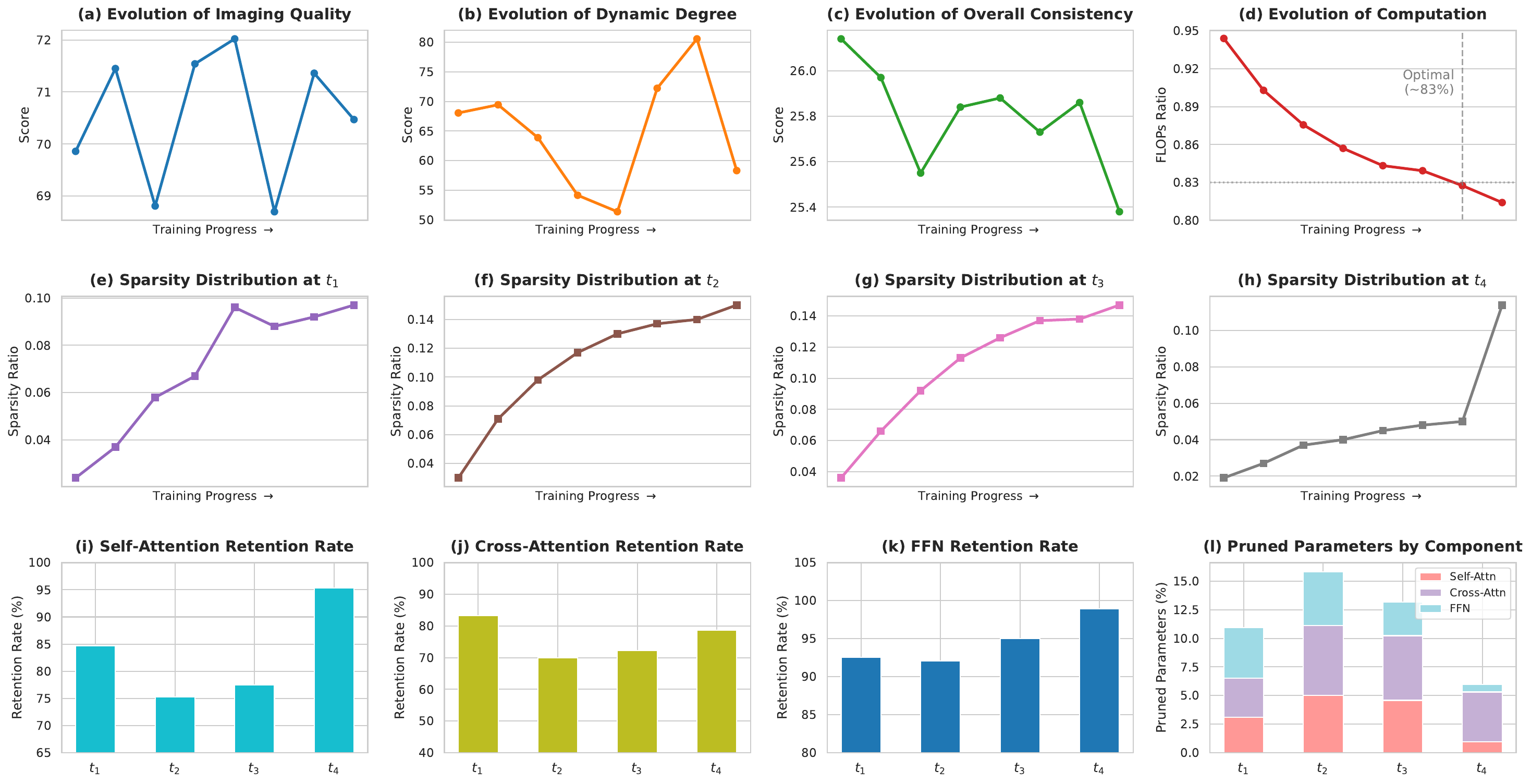}
    \caption{\textbf{Evolution of training dynamics and structural analysis.} (a)-(d) show the fluctuation of key generation metrics and the steady decrease in computational cost over training steps. (e)-(h) illustrate the increasing sparsity across four different denoising steps, $t_1$ presents the highest noise level. (i)-(l) provide a detailed component-level breakdown of the retained structures in our final selected model.}
    \label{fig:training_dynamics}
\end{figure}
\section{Extended Findings}
\label{sec:app_dynamics}
We track the evolution of the model's structural sparsity and its corresponding generation quality throughout the training process, as visualized in Figure~\ref{fig:training_dynamics} (on Wan 1.3B).

\textbf{Training Dynamics and the Sparsity-Quality Trade-off.}
We record the training dynamics starting from the Final State training. As the training progresses, the structural sparsity continuously increases across all denoising steps (Figure~\ref{fig:training_dynamics}(e)-(h)), leading to a steady decline in the global computation cost (Figure~\ref{fig:training_dynamics}(d)). However, thanks to our joint distillation strategy, the generation metrics do not degrade monotonically alongside the parameter reduction. Instead, the model maintains a highly competitive generation capability, with key metrics such as Imaging Quality and Dynamic Degree exhibiting stable fluctuations and even reaching peak performance at specific intermediate stages (Figure~\ref{fig:training_dynamics}(a)-(c)).

\textbf{Identifying a Favorable Sparsity-Quality Trade-off.}
Through this trajectory, we observe a practical capacity threshold that yields a highly favorable balance between computational efficiency and generation quality. The model demonstrates excellent performance when the average FLOPs ratio is reduced to approximately 83\%. At this specific operating point, the model attains a high Dynamic Degree and robust Imaging Quality. Pushing the sparsity beyond this threshold (e.g., dropping FLOPs below 81\%) results in a noticeable degradation, as excessively aggressive pruning begins to impair the representational capacity required for complex few-step generation. Consequently, we select the checkpoint at the $\sim$83\% FLOPs ratio as our final model, as it represents a well-balanced empirical sweet spot.

\textbf{Dynamic Degree Analysis on Wan 1.3B.}
Regarding the Dynamic Degree score of 80.56, which exceeds the teacher's 65.19, it is important to note two contextual factors. First, Dynamic Degree is an inherently high-variance metric. As shown in panels (a)--(c), it naturally fluctuates over $[51.4, 80.6]$ across viable training checkpoints. Second, While standard DMD degrades dynamics, our pipeline mitigates this via GAN training and extra real-video data: Ours~(Distill) alone already reaches 68.06($>$ teacher's 65.19). TurboDiffusion reaches 86.11, showing that a student exceeding its teacher on this metric is not anomalous.

Crucially, our method maintains an exceptionally high Motion Smoothness (98.17), which directly contradicts the possibility of temporal flickering or artefact-driven motion. Instead, we hypothesize that the structured stochastic perturbations introduced by our step-conditioned mask updates counteract distillation's mode-seeking tendency, yielding genuinely richer motion dynamics. This is corroborated by our supplementary videos, which exhibit larger, coherent motion rather than jitter.

\textbf{In-depth Structural Analysis.}
To further understand the resulting architecture of our joint optimization, we conduct a detailed component-level analysis, as shown in \cref{fig:training_dynamics}(i)-(l). The retention rates of Self-Attention, Cross-Attention, and Feed-Forward Network (FFN) modules exhibit distinct patterns across the denoising steps. As observed in \cref{fig:training_dynamics}(l), the sparsification of Self/Cross-Attention components constitutes a substantial portion of the total parameter reduction. The network learns to aggressively compress these Attention components particularly during the intermediate steps ($t_2$ and $t_3$), while maintaining a relatively higher retention rate for FFN modules. This fine-grained breakdown empirically illustrates how the joint optimization automatically distributes structural redundancy across different architectural components and inference stages.

\end{document}